\documentclass[runningheads]{llncs}
\usepackage[T1]{fontenc}
\usepackage{lipsum} 
\usepackage[disable]{todonotes} 

\usepackage{graphicx} 
\usepackage{subfigure}
\usepackage{tikz} 
\usepackage{adjustbox}  

\usepackage{booktabs} 
\usepackage{multirow} 
\usepackage{makecell} 
\usepackage{array, tabularx, boldline} 

\usepackage{amsmath} 

\usepackage{amsthm}
\usepackage{amsfonts}
\usepackage{bm} 
\usepackage{mathtools} 
\usepackage{mathdots} 
\usepackage{xfrac} 
\usepackage{faktor} 
\usepackage{cancel} 

\usepackage{algorithm}

\def\eqref#1{equation~\ref{#1}}

\def\1{\bm{1}}

\DeclareMathAlphabet{\mathsfit}{\encodingdefault}{\sfdefault}{m}{sl}
\SetMathAlphabet{\mathsfit}{bold}{\encodingdefault}{\sfdefault}{bx}{n}

\begin{document}
%
\title{Revisiting Bi-Encoder Neural Search: \\An Encoding--Searching Separation Perspective}  

%
\titlerunning{An Encoding--Searching Separation Perspective}

%
\author{Hung-Nghiep Tran\inst{1,2} \and
Akiko Aizawa\inst{3,4,5} \and
Atsuhiro Takasu\inst{3,4}}
%
%
\institute{University of Information Technology, Ho Chi Minh City, Vietnam \and
Vietnam National University, Ho Chi Minh City, Vietnam \and
National Institute of Informatics, Tokyo, Japan \and
The Graduate University for Advanced Studies, SOKENDAI, Tokyo, Japan \and
The University of Tokyo, Tokyo, Japan\\
\email{nghiepth@uit.edu.vn}, \email{\{aizawa,takasu\}@nii.ac.jp}}

\maketitle              

\begin{abstract}
	This paper reviews, analyzes, and proposes a new perspective on the bi-encoder architecture for neural search. While the bi-encoder architecture is widely used due to its simplicity and scalability at test time, it has some notable issues such as low performance on seen datasets and weak zero-shot performance on new datasets. In this paper, we analyze these issues and summarize two main critiques: the encoding information bottleneck problem and limitations of the basic assumption of embedding search. We then construct a thought experiment to logically analyze the encoding and searching operations and challenge the basic assumptions of embedding search. Building on these observations, we propose a new perspective on the bi-encoder architecture called the \textit{encoding--searching separation} perspective, which conceptually and practically separates the encoding and searching operations. This framework is applied to explain the root cause of existing issues and suggest mitigation strategies, potentially lowering training costs and improving retrieval performance. Finally, we discuss the broader implications of the ideas underlying this perspective, the new design surface it exposes, and potential research directions arising from it.

\end{abstract}

\section{Introduction}
Search is a crucial task in information retrieval that involves providing relevant results to user queries. Traditional lexical search methods have limitations in their ability to handle natural language queries to provide semantically accurate and relevant results \cite{robertson_probabilisticrelevanceframework_2009}. This has led to the emergence of neural search as a promising approach to search, offering several advantages over traditional methods, such as semantic matching ability \cite{lin_pretrainedtransformerstext_2021}.

One popular approach to neural search is the bi-encoder architecture, which has gained traction in recent years thanks to its simplicity and scalability at test time \cite{lin_pretrainedtransformerstext_2021}. This architecture involves separately encoding search queries and items as embedding vectors and then computing the similarity score between their embedding vectors. While this approach is promising, it also has some notable issues, including low results compared to other approaches such as cross-encoder \cite{urbanek_learningspeakact_2019}, low zero-shot results on new datasets \cite{thakur_beirheterogeneousbenchmark_2021}, and high training cost \cite{qu_rocketqaoptimizedtraining_2020}. These issues suggest that the bi-encoder architecture may have limitations that need to be addressed in order to improve its overall effectiveness in neural search. 

In this paper, we review and analyze these issues focusing on the two main critiques of the bi-encoder architecture: the \textit{encoding} information bottleneck problem and limitations of the basic assumption of embedding search, which is essentially \textit{encoding-for-search} \cite{lin_pretrainedtransformerstext_2021}. 
We conduct a thought experiment to logically analyze the encoding and searching operations in embedding search. The thought experiment takes the standard bi-encoder architecture and augments it with a searching operation, then we examine it under different scenarios to understand the roles of the encoding and the searching operations. The analyses indicate that the freedom between the encoding and searching operations is justifiable. These findings challenge the basic assumption of embedding search. We argue that \textit{encoding-for-search} is not always necessary, but rather a design choice.

Based on our observations, we propose the novel \textit{encoding--searching separation} perspective on the bi-encoder architecture. This perspective offers a conceptual and practical separation of the encoding and searching operations in embedding search. The induced augmented model has a searching operation on top of the query encoding operation to create a flexible ``\textit{encoding gap}'', which is the difference of information between the encoding and searching tasks. 

This perspective has several advantages, including better control of the information bottleneck, greater freedom in designing the encoding and searching operations, and improved training efficiency. Although mathematically equivalent to the standard bi-encoder architecture, the new perspective represents a significant shift in the underlying ideas, which enables us to better understand and naturally modify the bi-encoder architecture to mitigate its problems.

Overall, this paper contributes to the understanding of the bi-encoder architecture and provides new insights for potentially improving its performance in neural search. 
Our contributions are as follows: 
\begin{itemize}
	\item We review and analyze some important issues of the bi-encoder architecture based on two main critiques: the encoding information bottleneck problem and limitations of the basic assumption of embedding search.
	
	\item Through a thought experiment, we propose the new \textit{encoding--searching separation} perspective on the bi-encoder architecture. This perspective challenges the basic assumption of embedding search and provides a new approach to understanding and implementing the bi-encoder architecture.
	
	\item Using the proposed perspective, we analyze the root causes of the identified issues of the bi-encoder architecture and suggest potential mitigations.
	
	\item Finally, we discuss the implications of the proposed perspective, including new design surfaces that it exposes and potential research directions.
\end{itemize}

\section{Background}
\subsection{Search Task}

The search task is the process of finding relevant information from a large collection of items based on a user’s query. This involves matching the query with the items and presenting the most relevant results to the user in a ranked order \cite{lin_pretrainedtransformerstext_2021}. 
Recently, zero-shot search has become an increasingly important task, where no training data is available to train or fine-tune the system. As a result, such search systems requires generalization to out-of-distribution data \cite{thakur_beirheterogeneousbenchmark_2021}.

\subsection{Search Approaches} 
The search task can be performed using various techniques and algorithms, ranging from lexical keyword-based search to more advanced semantic methods. 

The traditional lexical full-text search approach matches a search query against the full text of search items. 
Traditional full-text search is widely used in various applications. However, it has some limitations, such as the lack of semantic understanding and the inability to handle complex queries \cite{mitra_introductionneuralinformation_2018}.

Neural search is a technique that train a neural network to learn the relationship between queries and search items. Neural search has been promising in overcoming some the limitations of traditional search, such as semantic understanding, the ability to handle natural language queries, and contextual understanding \cite{mitra_introductionneuralinformation_2018}. 
In recent years, an emerging trend in neural search is models utilizing large pretrained transformers models \cite{lin_pretrainedtransformerstext_2021}. These models are built upon the ability to learn meaningful contextual embedding for natural language text of transformers such as BERT \cite{devlin_bertpretrainingdeep_2018}. There are three main architectures:

\begin{itemize} 
	\item Cross-encoder: Jointly encodes the concatenation of \textit{query} and \textit{item} as one vector, then uses it as features for a scoring function to compute relevance score. This approach achieves good results but has a high cost \cite{nogueira_passagererankingbert_2019} \cite{bajaj_msmarcohuman_2018}.
	
	\item Bi-encoder: Separately encodes the \textit{query} and \textit{item} as two embedding vectors, then computes their vector similarity to measure relevance score. This approach is simple and scalable, as the embeddings for the items can be precomputed and indexed for efficient search \cite{lin_pretrainedtransformerstext_2021}. 
	
	\item Poly-encoder: This can be seen as a variant of the bi-encoder architecture, where the similarity function is a complex attention-based model and may require computation for each item one-by-one \cite{humeau_polyencodersarchitecturespretraining_2020}.
\end{itemize}

\subsection{Bi-Encoder Architecture}
We mainly focus on the bi-encoder architecture because of its simplicity and scalability. In the most general form, the bi-encoder architecture aims to produce embeddings (i.e., vector representations) of queries and items that are similar to each other if they are relevant to each other and vice versa \cite{lin_pretrainedtransformerstext_2021}. 

The relevance score of the general bi-encoder architecture is computed by:
\begin{align}
	P(rel = 1 | q, d) = s\Big(sim\big(enc1(q), enc2(d)\big)\Big), \label{eq:biencoder-general}
\end{align}
where $P$ is the probability that the item $d$ is relevant to the query $q$. The function $s(\cdot)$, usually sigmoid, is to compute the probability. $enc1(\cdot)$ and $enc2(\cdot)$ denote the query and item encoders, respectively. The similarity function $sim(\cdot, \cdot)$ computes the similarity score between the query and the item embeddings.

In practice, the function $sim(\cdot, \cdot)$ must be fast using existing hardware and algorithms. One of such functions is the inner product, where the item embedding can be precomputed and indexed for fast approximate nearest neighbor search \cite{matsui_surveyproductquantization_2018}, which can scale to billions of items \cite{johnson_billionscalesimilaritysearch_2017}. The relevance score of a practical bi-encoder architecture is rewritten as:
\begin{align}
	P(rel = 1 | q, d) = s\Big(\big\langle enc1(q), enc2(d) \big\rangle \Big), \label{eq:biencoder-dotproduct}
\end{align}
where $\langle \cdot, \cdot \rangle$ denotes the inner product between the embedding vectors. Note that the inner product is fast not just because it is simple, but because the item embeddings can be indexed in advance, instead of requiring computation one-by-one as in other approaches.

\section{Critiques of the Bi-Encoder Architecture}
The bi-encoder architecture is a promising approach with several good properties, for example simple architecture, precomputed item embeddings, fast vector search algorithms. However, the bi-encoder architecture also has several issues: 

\begin{enumerate}
	\item \textit{Low results:} Given similar training data, the bi-encoder architecture achieves lower results than other approaches such as cross-encoder \cite{karpukhin_densepassageretrieval_2020} \cite{nogueira_passagererankingbert_2019}. 
	
	\item \textit{Low zero-shot results:} Most bi-encoder models can be fine-tuned to get very good results on the MSMARCO dataset, but on new datasets in the BEIR zero-shot benchmark, they get even lower results than BM25 \cite{thakur_beirheterogeneousbenchmark_2021}. 
	
	\item \textit{Expensive fine-tuning cost:} This issue is often overlooked because the bi-encoder architecture is fast at test time. However, at training time, it is almost as expensive as the cross-encoder architecture, because each query and item needs to be re-encoded to train with each sample \cite{karpukhin_densepassageretrieval_2020} \cite{qu_rocketqaoptimizedtraining_2020}. 
	
	\item \textit{Tendency to overfit:} This is the underlying technical issue when training the bi-encoder architecture. For example, the bi-encoder DPR model \cite{karpukhin_densepassageretrieval_2020} has both low in-domain results on MSMARCO and low zero-shot results on BEIR \cite{thakur_beirheterogeneousbenchmark_2021}.
\end{enumerate}

\subsection{Analyses and Main Critiques of Bi-Encoder Architecture}
Here we analyze the above issues. Our analyses focus on the two main critiques: the encoding information bottleneck problem and limitations of the encoding-for-search assumption.

\subsubsection{The Encoding Information Bottleneck Problem}
In the context of the bi-encoder architecture, it is a popular belief that the fixed-size embeddings become an information bottleneck that restricts the expressiveness of the model \cite{khattab_colbertefficienteffective_2020}. However, we note that real-value embedding can carry almost infinite amount of information, up to the numerical limit. Thus, we argue that the bottleneck is actually induced by the encoding process that drops useful information, rather than the embedding. We call this the \textit{encoding} information bottleneck.

About issues (1) and (2), when a bi-encoder model achieves low results, previous work usually try to widen the information bottleneck by using larger embedding or multiple embeddings \cite{khattab_colbertefficienteffective_2020}. In stead, the \textit{encoding} information bottleneck suggests that we should use a larger encoder. Moreover, we should aim to separate the encoding process from the search task, which means better control of the information bottleneck.

So, the first critique of the bi-encoder architecture is: \textit{the encoding information bottleneck is spreading throughout the encoder}. Overall, we argue that better control of the information bottleneck is important in improving the retrieval performance of the bi-encoder architecture. 

\subsubsection{Limitations of the Basic Assumption of Embedding Search}
The basic assumption of embedding search is that the relevance score of a query and a search item can be computed by measuring the similarity between their embeddings \cite{lin_pretrainedtransformerstext_2021}. This implies that the embeddings must capture the information needed for the search task. As a result, the encoders directly serve the search task. We call this the encoding-for-search assumption.

About issues (3) and (4), the cost of fine-tuning is high because when fine-tuning for search on a new dataset, we have to fine-tune the whole encoder for this search task, which may be unnecessarily expensive. 
In addition, fine-tuning the whole encoder tends to overfit because the search task is relatively simple compared to the capacity of large transformer-based encoders.

Furthermore, in the case of the practical bi-encoder architecture, in Eq. \ref{eq:biencoder-dotproduct}, the inner product implies that the embeddings of query and item must be close to each other in the embedding space. While this assumption may seem reasonable for document retrieval, it becomes challenging in the context of multi-modal data such as videos or music, where different modalities operate in separate embedding spaces. In such cases, forcing the alignment of the embedding spaces across different modalities can be unnatural and suboptimal for retrieval performance. 

So, the second critique of the bi-encoder architecture is: \textit{the encoding-for-search assumption may be too strong}. We argue that this assumption should be challenged and modified to improve the bi-encoder architecture.

\section{New Perspective on the Bi-Encoder Architecture}
In this section, we conduct a thought experiment to analyze the roles of encoding and searching, and challenge the \textit{encoding-for-search assumption}. We then define the \textit{encoding--searching separation perspective} and discuss its implications.

\begin{figure*}[ht]
	\centering
	\centerline{\includegraphics[width=1.2\textwidth]{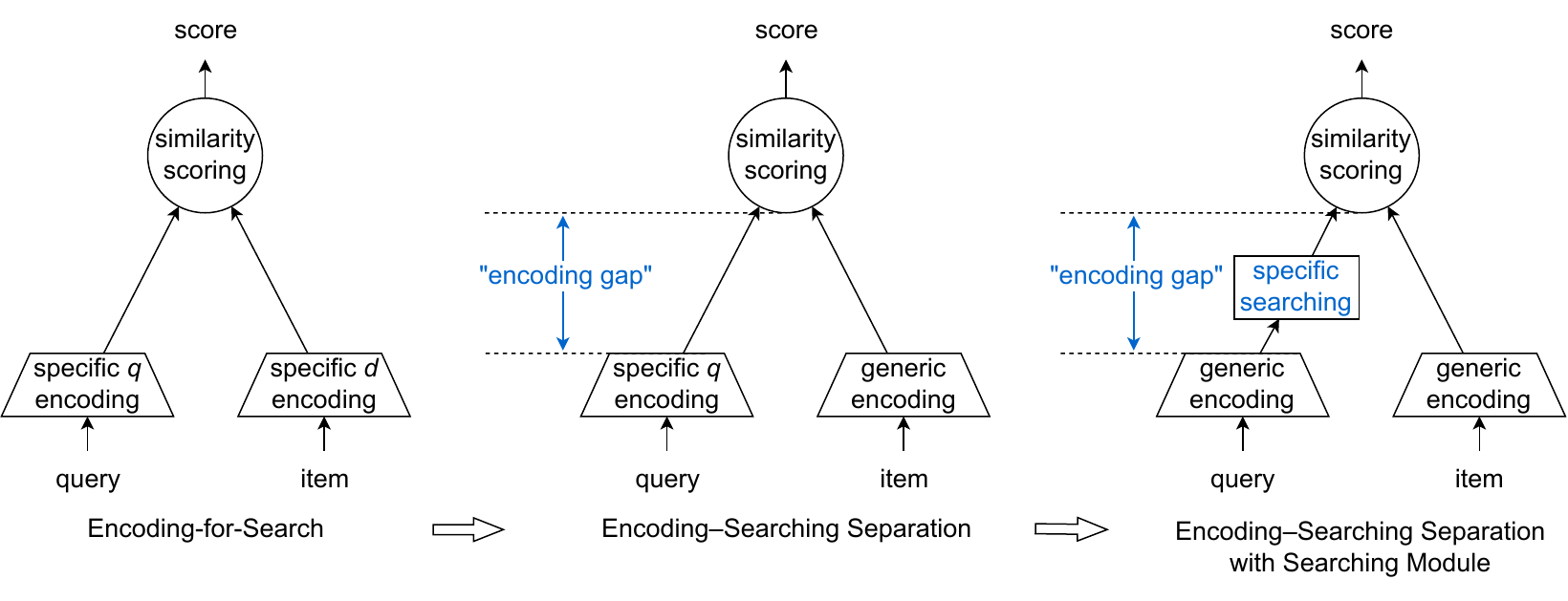}}
	\caption{Illustration of the Encoding--Searching Separation perspective and the induced ``encoding gap'', which is the difference of the information between the encoding and searching tasks. Note the change from \textit{specific encoding} to \textit{generic encoding}.}
	\label{fig:encoding-searching-separation}
\end{figure*}

\subsection{Thought Experiment}
The thought experiment is conducted as follows. Given the bi-encoder architecture, we add a transformation on top of the query embedding, then compute the similarity score between its output with the item embedding as normal. The encoder is now called the \textit{encoding module} performing the \textit{encoding operation}. The added transformation is called the \textit{searching module} performing the \textit{searching operation}. We analyze the behavior of the modified architecture under different scenarios to explore the properties of the encoding and the searching operations.

\begin{itemize}
	\item Scenario 1: Assume that the encoding operation encodes any query on any dataset as vector zero. This means the encoding operation provides no information. In terms of information theory, it removes all information present in the queries (or any input), the entropy is reduced to zero. 
	\begin{itemize}
		\item In this case, there is no information to differentiate between different queries and to find their relevant items. Thus, even with the best searching operation, there is no way the model could find relevant items. The model fails on all datasets. 
		
		\item Hence, a necessary condition for the model to work is that the encoding operation must provide information. 
	\end{itemize}
	
	\item Scenario 2: Assume that the searching operation maps all embeddings to vector zero. This means the searching operation ignores all information in the embedding. 
	\begin{itemize}
		\item In this case, even with the best encoding operation that provides useful information, there is no way the model could find relevant items. The model fails on all datasets. 
		
		\item Hence, a necessary condition for the model to work is that the searching operation must select and retain some information. 
	\end{itemize}
	
	\item Scenario 3: Assume that a frozen encoding operation provides the exactly needed information for a search task on a given dataset, but no information for other datasets. 
	\begin{itemize}
		\item When fine-tuning on the given dataset, the searching operation can easily select and retain the needed information for the search task. The model can work well on the given dataset. 
		
		\item When fine-tuning on a new dataset, there is no useful information for the searching operation to select from. Thus, even with the best searching operation, the model fails on new datasets. 
		
		\item In this case, the encoding operation becomes an information bottleneck because it is too specific for a search task. Hence, the encoding operation should be generic. 
	\end{itemize}
	
	\item Scenario 4: Assume that the encoding operation provides rich and generic information for all search tasks on all datasets. 
	\begin{itemize}
		\item When fine-tuning on the given dataset, the searching operation can select and retain the needed information for the search task. The model can work well on the given dataset. 
		
		\item When fine-tuning on a new dataset, the searching operation can also select and retain the needed information for the new search task. The model can work well on new datasets. 
		
		\item In this case, the model can work because the searching operation can be fine-tuned to select new useful information from the generic embedding. Otherwise, if the searching operation is frozen, it will become an information bottleneck. Hence, we are able to localize the information bottleneck in the model.
	\end{itemize}
\end{itemize}

By augmenting the bi-encoder architecture with a search transformation, we can investigate the encoding and the searching operations separately. From scenarios 1 and 2, we see that the encoding and searching operations have different roles. The encoding operation provides information, whereas the searching operation selects and retains information. From scenario 3, we see that the encoding operation may become problematic if it is specific to a search task. To avoid this problem, the encoding operation should be generic. Overall, we see that \textit{encoding-for-search} is not necessary because encoding and searching can be separated into two different operations. Moreover, \textit{encoding-for-search} is a too strong assumption that may harm the model performance because it makes the encoding operation become too specific for a search task.

From scenario 4, we see that a generic encoding operation and a specific searching operation can work well together. Thus, a different design choice than the \textit{encoding-for-search} is the \textit{generic encoding--specific searching} approach. In this approach, the encoding operation is task-agnostic and its role is to provide generic information. The searching operation is task-specific and its role is to select and compose specific information that is useful for the search task.

\subsection{The Encoding--Searching Separation Perspective}
The main ideas of the \textit{encoding--searching separation perspective} is to conceptually and practically separate the encoding and searching operations in the bi-encoder architecture. Figure \ref{fig:encoding-searching-separation} illustrates an encoding--searching separation model with the ``encoding gap'' between the encoding and searching tasks. The relevance score of a practical \textit{encoding--searching separation model} is written as:
\begin{align}
	P(rel = 1 | q, d) = s\Big(\big\langle f_{search}\big(enc1'(q)\big), enc2(d) \big\rangle \Big), \label{eq:biencoder-dotproduct-separation}
\end{align}
where $f_{search}(\cdot)$ is the specific searching module on top of the generic encoding module $enc1'(\cdot)$. Note that we will use the terms encoding and searching \textit{operations} conceptually, and the terms encoding and searching \textit{modules} practically in a specific model. 

In details, the encoding modules $enc1'(\cdot)$ and $enc2(\cdot)$ provide generic embeddings for any search task on any dataset. The search module $f_{search}(\cdot)$ does the heavy-lifting for each specific search task by selecting and composing specific information from the query embedding that is useful for the search task. 

The objective of the augmented \textit{encoding--searching separation model} remains the same as the original bi-encoder architecture, which is to assign high similarity scores to relevant items. However, the similarity score is no longer computed directly from the encoding output, but rather from the searching function output, $f_{search}\big(enc1'(\cdot)\big)$. Therefore, there exists an ``encoding gap'' between the encoding operation and the search task that gives us more flexibility and control in the model. The searching operation can be viewed as \textit{selective} embedding alignment that fills the ``encoding gap''. This aligns the useful features between the query encoding output and the item encoding output, with the advantage of being more efficient and easier to train than aligning the whole embedding spaces.

\section{Discussions}
Here we summarize the main implications, revisit and mitigate the identified issues, and explore new research directions.

\subsection{Main Implications} 

The encoder in the bi-encoder architecture can be viewed as composing of the encoding and searching operations. These operations have different roles and should be treated accordingly. The encoding operation is generic and can be task-agnostic. The searching operation must be specific for the search task. 

In an \textit{encoding--searching separation} model, we can control the location of the information bottleneck by moving it between the encoding and the searching operations. By using generic encoding operation, the information bottleneck moves from the encoding to searching operation and it becomes easier to transfer the generic encoding operation to a new search task on other datasets. We may also aim to widen the information bottleneck in the model by exploring different designs and training strategies for the searching modules. 

There is an ``encoding gap'' between the encoding operation and the search task, which would be filled by the searching operation. This gap enables a \textit{generic encoding--specific searching} setting, in which the encoding operation is free from fine-tuning requirement, whereas the searching operation does the heavy-lifting for the search task by selecting and composing useful information from the generic encoding output. 
The searching operation can be seen as a form of selective embedding alignment, which aligns useful embedding features between the query encoding output and the item encoding output, which is more efficient than aligning the whole embedding spaces. Furthermore, we can use a frozen generic encoding operation or even precomputed embeddings to train the searching operation alone. In this case, because embeddings are not computed online, we may use exceptionally high settings to train the searching operation, such as very large batch sizes and expensive loss functions.

\subsection{Revisit the Issues and Critiques of the Bi-encoder Architecture}
Using the new perspective, we try to explain the cause of the identified issues and suggest ways to mitigate them.

\subsubsection{Root Cause of the Identified Issues} 
We argue that the identified issues are the results of a \textit{failure to separate the encoding and searching operations} in the bi-encoder architecture due to the \textit{encoding-for-search} assumption. Firstly, about issue (3), fine-tuning the whole encoder is unnecessarily expensive, because both the encoding and searching operations are fine-tuned for a specific search task, whereas we may fine-tune the searching operation only. Secondly, about issue (4), fine-tuning both the encoding and searching operations for a specific search task is more prone to overfitting, because both the encoding and searching operations contain the information bottlenecks. Thirdly, about issue (2), when both encoding and searching are fine-tuned for a specific search task, encoding becomes specific and searching is fine-tuned for such specific encoding output. No part of the model is general and easily transferable to other datasets. Finally, about issue (1), the encoding module usually has rich capacity, so it may be fine-tuned to become too specific and provides low quality embeddings for the searching operation, for example, spurious features instead of semantic features. This causes severe overfitting and may lead to low result even on a single dataset.

\subsubsection{Potential Mitigations for the Critiques}
The new perspective suggests some simple solutions to the problems in the two critiques. 
\begin{itemize}
	\item For the \textit{encoding information bottleneck problem}, we can gain better control by localizing the bottleneck to a specific part of the model. For instance, instead of dealing with bottlenecks on both the encoding and searching modules, we can move it to the searching module by learning a specific searching operation on top of a generic encoding operation. When fine-tuning the searching operation, it is less prone to overfitting because it is forced to work with generic encoding output. In addition, the generic encoding operation may be easily transferred to a new search task on other datasets. 

	\item For \textit{limitations of the encoding-for-search assumption}, we can conceptually separate the encoding and searching operations to create a flexible ``encoding gap'' between the encoding operation and the search task. This encoding gap relaxes the fine-tuning requirements on the encoding operation and lessens the influence of the encoding-for-search assumption.
\end{itemize}

\subsection{New Research Directions} 
The new perspective highlights the critical relationship between the encoding and searching operations, revealing an understudied design surface that offers a range of configurations and design choices to potentially improve the efficiency and effectiveness of the bi-encoder architecture for neural search. Here we identify some important research directions that may benefit the research community.

\begin{itemize}
	\item \textit{Fixed Encoding, Trained Searching:}
	This direction is motivated by the observation that the encoding and searching operations have different roles, and asks how much these operations can be separated. We consider the case that the encoding is fixed and only the searching operation is trained for a search task, so that the searching operation does all the heavy-lifting in the search task. This is an important case because it may enable the use of frozen encoding operations, allowing the searching operation to be trained separately on precomputed embeddings, even using exceptionally high settings. 
	
	\item \textit{Investigate the Information Bottleneck:}
	This research direction seeks to address the information bottleneck problem in the bi-encoder architecture when a single embedding is used. By separating the encoding and searching operations, it is possible to localize and analyze the bottlenecks explicitly. 
	
	\item \textit{Design and Train Better Searching Operations:}
	This research direction aims to investigate the potential design and training strategy of searching operations that can perform the heavy-lifting in the search task while being more efficient than the traditional bi-encoder approach. 

	\item \textit{Design Customized Models for Transfer Learning:}
	This research direction aims to explore new configurations enabled by the encoding--searching separation perspective that are suitable for zero-shot search and transfer learning between search tasks, an important topic in the field of neural search. 
\end{itemize}

\section{Related Work}
There have been several surveys on neural search such as \cite{mitra_introductionneuralinformation_2018} providing an introduction to neural search, \cite{lin_pretrainedtransformerstext_2021} describing the emerging trend of using pretrained transformers for neural search. However, these surveys do not specifically review and analyze the identified issues of the bi-encoder architecture in this paper. The issue of low zero-shot search result has been discussed extensively in \cite{thakur_beirheterogeneousbenchmark_2021}, which inspired us to look deeper into the problems and its root cause. The proposed perspective provides an explanation to this issue and suggests potential solutions. 

Bi-encoder architecture is a popular and promising approach to neural search. There are many popular pretrained transformer-based models, such as DPR \cite{karpukhin_densepassageretrieval_2020} and RocketQA \cite{qu_rocketqaoptimizedtraining_2020}. These models all use the standard bi-encoder architecture with two encoders for the query and the item, respectively, and an inner product similarity function. They mainly differ in their training strategy such as hard negative sampling and distilling from cross-encoder model. 

There are several extensions of the bi-encoder architecture such as ColBERT \cite{khattab_colbertefficienteffective_2020} that uses all tokens’ embeddings as encoding output and a specialized similarity function, Poly-encoder \cite{humeau_polyencodersarchitecturespretraining_2020} that generates multiple embeddings for the query and the item, then uses a complex similarity function to search, which is costly at test time, because the search items need to be accessed one-by-one. These works have some similarity to the proposed perspective in that they have some functions on top of the encoding output. However, they do not explicitly view the model in the perspective of \textit{encoding--searching separation}. In fact, our approach may be used to explain some properties of these models. 

Our approach shares some similarities with some old neural search methods predating transformers that work on fixed word embeddings \cite{mitra_introductionneuralinformation_2018}. However, they did not have access to context-rich generic encoding operation such as pretrained large language models, as a result, their perspectives did not consider the generic embedding as an advantage, and usually focused on developing complex similarity functions. Instead, we extensively analyze the \textit{generic encoding--specific searching} setting to gain better understandings of modern bi-encoder neural search and explore potential improving directions. 

In other domains, ideas and approaches sharing some similarities to this perspective have shown good results. For example, in knowledge graph embedding, the relational transformation can be separated from the entity encoding operation, enabling expressive relational modeling and efficient training \cite{tran_multipartitionembeddinginteraction_2020} \cite{tran_meimmultipartitionembedding_2022}. In data analysis, rich semantic queries may be performed directly on the fixed embedding space \cite{tran_multirelationalembeddingknowledge_2020} \cite{tran_exploringscholarlydata_2019}. In a similar way, the proposed perspective may contribute to both conceptual understanding and better results for neural search.

\section{Conclusion}
In this paper, we analyzed and summarized the main critiques of the bi-encoder architecture, including the \textit{encoding} information bottleneck problem and limitations of the basic assumption of embedding search. Through a thought experiment, we challenged the basic assumption of embedding search and proposed the new \textit{encoding--searching separation} perspective, which conceptually and practically separate the encoding and the searching operations. Based on the new perspective, we explained the root causes of the identified issues of the bi-encoder architecture and explored potential mitigations. Finally, we discussed the broader implications of the new perspective, the design surface it exposes, and potential research directions arising from it.

While this paper introduces a new perspective on bi-encoder architectures and provides theoretical insights, it currently lacks extensive empirical experiments across diverse datasets. Demonstrating practical advantages, such as improved retrieval performance and reduced training costs, requires rigorous empirical evaluation. To address this, we plan to conduct experiments using MS MARCO \cite{bajaj_msmarcohuman_2018} for supervised training and BEIR \cite{thakur_beirheterogeneousbenchmark_2021} for zero-shot evaluation, comparing our approach with standard bi-encoders in terms of accuracy, computational efficiency, and the trade-off between accuracy and compute. A particularly promising direction is training an ensemble of lightweight searching modules on top of precomputed frozen embeddings, enabling fast and efficient training. Investigating how far these modules can bridge the encoding gap and how their composition contributes to generalization across different zero-shot datasets would provide valuable insights. We hope this work lays the foundation for future research, inviting further exploration into the interplay between encoding and searching to advance neural retrieval models.


%
%
%
\bibliographystyle{splncs04}

\appendix
\section*{Appendix}

\section{Remarks on the Proposed Perspective}
In this paper, we derive the augmented model in Eq. \ref{eq:biencoder-dotproduct-separation} from the proposed perspective and the practical bi-encoder architecture in Eq. \ref{eq:biencoder-dotproduct}. This model uses the inner product as the similarity function, enabling efficient indexing and searching. A similar derivation can be applied to the general bi-encoder architecture in Eq. \ref{eq:biencoder-general}, but it is omitted for simplicity because its complex similarity functions may be impractically expensive. 

\subsection{On Equivalence and Difference} 
Note that the \textit{encoding--searching separation perspective} still describes the original practical bi-encoder architecture in Eq. \ref{eq:biencoder-dotproduct}. In fact, we can show that they are mathematically equivalent as follows. 

On one hand, we can merge two functions, $enc1'(\cdot)$ and $f_{search}(\cdot)$, in an arbitrary augmented \textit{encoding--searching separation model} into a new encoder:
\begin{align}
	enc1''(\cdot) =\ &f_{search} \circ enc1'(\cdot) \nonumber \\
	=\ &f_{search}\big(enc1'(q)\big), \label{eq:biencoder-separation-equivalence}
\end{align}
and use it in a model of original bi-encoder architecture. On the other hand, we can split the encoder $enc1(\cdot)$ in a certain model of original bi-encoder architecture into two functions, $enc1'(\cdot)$ and $f_{search}(\cdot)$, and use them in an augmented model. 

Certain extensions of the bi-encoder architecture, which use more complex similarity functions, such as the poly-encoder model \cite{humeau_polyencodersarchitecturespretraining_2020} and ColBERT \cite{khattab_colbertefficienteffective_2020}, may look similar to the proposed perspective but are distinct from it. Firstly, practically, these models use multiple embeddings, and their complex similarity functions may require the item embeddings to be accessed one-by-one at test time, as is the case with poly-encoder. Secondly, and more importantly, conceptually, they do not separate the searching from the encoding operation and are thus still limited by the encoding-for-search assumption.

\subsection{On Meaning and Contribution}
Two models that are mathematically equivalent can be meaningfully different if the ideas behind them are distinct and significant. This is because although their current results and predictions are the same, their distinct ideas can enable different insights, leading to useful modifications that are natural in one model but not in the other. This pattern is common in experimental sciences like physics and computer science.

This paper presents an initial exploration of the understandings and possibilities enabled by the ideas behind the new perspective. Although the proposed \textit{encoding--searching separation} perspective is simple, it is useful in explaining the identified issues, suggesting mitigations, opening up a broader surface for model design, and guiding natural research directions.

\section{Remarks on the New Research Directions}  
\subsection{Fixed Encoding, Trained Searching:}
At first glance, this direction may seem infeasible. However, it makes sense if we think of generic embedding as a high-fidelity representation suitable for different tasks. In fact, the emerging trend in NLP is to use large pretrained models \cite{devlin_bertpretrainingdeep_2018} \cite{brown_languagemodelsare_2020}, which makes fixed encoding more natural or even required. Moreover, recent research has shown that fine-tuning transformers for the classification task keeps the linguistic features mostly unchanged, with most changes occurring in the classifier \cite{merchant_whathappensbert_2020}, which provides a technical hint for this direction. 

Research in this direction should answer how much the searching operation alone can do the heavy-lifting in the search task, and how to leverage new architectures and training strategies to train a good searching operation.

\subsection{Investigate the Information Bottleneck:}
To widen the bottleneck with a single embedding, the key is to have a good encoding operation. One possibility is to leverage pretrained large language models such as GPT which have been shown to be strong multi-task learners \cite{radford_improvinglanguageunderstanding_2018} \cite{radford_languagemodelsare_2019} \cite{brown_languagemodelsare_2020}. We need to study methods to obtain generic embedding from such models and verify whether these embeddings are generic enough for different search tasks. 

Another important topic is exploring the interplay between the generic encoding operation and the specific searching operation. We may look into possibilities to widen the information bottleneck via special training strategies.

\subsection{Design and Train Better Searching Operations:}
There are various architectures that can be explored, from simple linear layers to large transformer models. In addition, we may design new architectures specifically suited for the searching operation inspired by recent understanding of memory stored in transformers \cite{meng_locatingeditingfactual_2022}. 
	
Another interesting topic is training strategy for the searching operation. Here we present a special case to show how it can improve the performance at a low cost. First, by initializing the encoding module $enc1'(\cdot)$ using an already fine-tuned encoder $enc1(\cdot)$ and fixing it, we can create an augmented model that is at least as good as the original bi-encoder model with $enc1(\cdot)$ on a given dataset. Then, we can use precomputed embeddings to efficiently train $f_{search}(\cdot)$ to further enhance the retrieval performance on this dataset.
	
\subsection{Design Customized Models for Transfer Learning:}
By separating the encoding and searching operations, one obvious solution is by transferring the generic encoding operation to new search tasks. However, we may also consider how to transfer the searching operation. In case the two search tasks are similar, it makes sense to transfer the searching operation directly. But in cases where the two tasks have both similarities and dissimilarities, we need to be able to transfer only the similar parts of the searching operation. 

This direction can be seen as a generalization of the \textit{encoding--searching separation} perspective to a fully modular architecture, in which the encoding and searching operations are further separated into sub-operations, some of which are transferable. This modular architecture approach may open a new area for designing and training effective models.

\end{document}